\documentclass[journal,10pt,twocolumn]{IEEETran}
\usepackage[utf8]{inputenc}

\usepackage{threeparttable}
\usepackage{cite}
\usepackage{amsmath,amssymb,amsfonts}
\usepackage{algorithmic}
\usepackage{graphicx}
\usepackage{booktabs}
\usepackage{textcomp}
\usepackage{indentfirst}
\usepackage{graphicx}
\usepackage{csquotes}
\usepackage{multirow}
\usepackage{tabularx}
\usepackage{float}
\usepackage{color}
 \usepackage{url}
\usepackage{framed}
\usepackage{comment}
\usepackage{caption2}
\usepackage{algorithm2e,setspace}
\usepackage{color, colortbl}
\usepackage{hhline}
\usepackage{graphicx,subfigure}

\SetKwProg{Fn}{Function}{}{}
\definecolor{Gray}{gray}{0.9}
\newcolumntype{g}{>{\columncolor{Gray}}c}

\usepackage{bbding}

\makeatletter
\DeclareRobustCommand\onedot{\futurelet\@let@token\@onedot}
\def\@onedot{\ifx\@let@token.\else.\null\fi\xspace}

\def\ie{\emph{i.e}\onedot}

\makeatother

\renewcommand{\figurename}{Figure}
\def\tablename{Table}

\begin{document}

\title{The art of defense: letting networks fool the attacker}

\author{Jinlai Zhang, Yinpeng Dong, Binbin Liu, Bo Ouyang, Jihong Zhu, Minchi Kuang, Houqing Wang, Yanmei Meng\\

\thanks{Corresponding author: Jihong Zhu. Jinlai Zhang, Yanmei Meng are with the College of Mechanical Engineering, Guangxi University, Nanning, China (email: cuge1995@gmail.com, gxu\_mengyun@163.com). Yinpeng Dong, Binbin Liu, and Bo Ouyang are with the Department of Computer Science and Technology, Tsinghua University, Beijing, China (email: dongyinpeng@gmail.com, lbb19@mails.tsinghua.edu.cn, oyb19@mails.tsinghua.edu.cn). Jihong Zhu, Minchi Kuang, and Houqing Wang are with the Department of Precision Instrument, Tsinghua University, Beijing, China (email: kuangmc@mail.tsinghua.edu.cn, houqingok@163.com, jihong\_zhu@hotmail.com).}} 

\maketitle

\begin{abstract}
Robust environment perception is critical for autonomous cars, and adversarial defenses are the most effective and widely studied ways to improve the robustness of environment perception.
However, all of previous defense methods decrease the natural accuracy, and the nature of the DNNs itself has been overlooked. To this end, in this paper, we propose a novel adversarial defense for 3D point cloud classifier that makes full use of the nature of the DNNs. Due to the disorder of point cloud, all point cloud classifiers have the property of permutation invariant to the input point cloud. Based on this nature, we design invariant transformations defense (IT-Defense).
We show that, even after accounting for obfuscated gradients, our IT-Defense is a resilient defense against state-of-the-art (SOTA) 3D attacks. Moreover, IT-Defense do not hurt clean accuracy compared to previous SOTA 3D defenses. Our code will be available at: {\url{https://github.com/cuge1995/IT-Defense}}.
\end{abstract}


\section{Introduction}
Deep neural networks (DNN) has shown great success in many fields~\cite{yolo,gpt3,pointnet++,widerresnet,heresnet,monitoringsugar}. However, they are vulnerable to maliciously generated adversarial examples~\cite{xu2020adversarial}. As DNN models have been implemented into various real-world applications, \ie, face recognition~\cite{face} and autonomous driving~\cite{pvrcnn,xu2020adversarial}, the research on adversarial robustness has attracted more and more attention, and many adversarial attack algorithms have been proposed, this puts many DNN models deployed in the real world under serious threats. Therefore, it is crucial to conduct extensive research on adversarial defense.

Adversarial training is considered to be the most effective defense and it can generalize across different threat models~\cite{benchmarkingimage}. However, adversarial training faces many problems. Firstly, the high cost of standard adversarial training making it impractical. In order to reduce the cost of standard adversarial training, Shafahi et al. \cite{shafahi2019adversarialadvt1} recycled the gradient information that was computed when updating model parameters, it finally speed up adversarial training 7 to 30 times compared to the standard adversarial training. Maksym et al. \cite{andriushchenko2020understandingadvt2} proposed the GradAlign that that prevents catastrophic overfitting in fast gradient sign method (FGSM) training. Secondly, all adversarial training methods cannot overcome the problem that adversarial trained model decrease the recognition accuracy in clean samples. Another promising line of defense against adversarial examples is to randomize the inputs or the model parameters \cite{pinot2020randomization}. The randomize of the model parameters is to sample weights of networks from some distribution \cite{liu2018towards, liu2018adv, lee2022graddiv}, however, it needs to adapted with the target networks. The random transforms to the input in 2D image as defense methods \cite{zhang2019defending, raff2019barrage, inputtrans,cohen2019certified} has been studied extensively, and have shown excellent robustness, but it rarely explored in 3D point cloud. In this paper, we focus on input random transforms in 3D point cloud.



\begin{figure}
  \includegraphics[width=\linewidth]{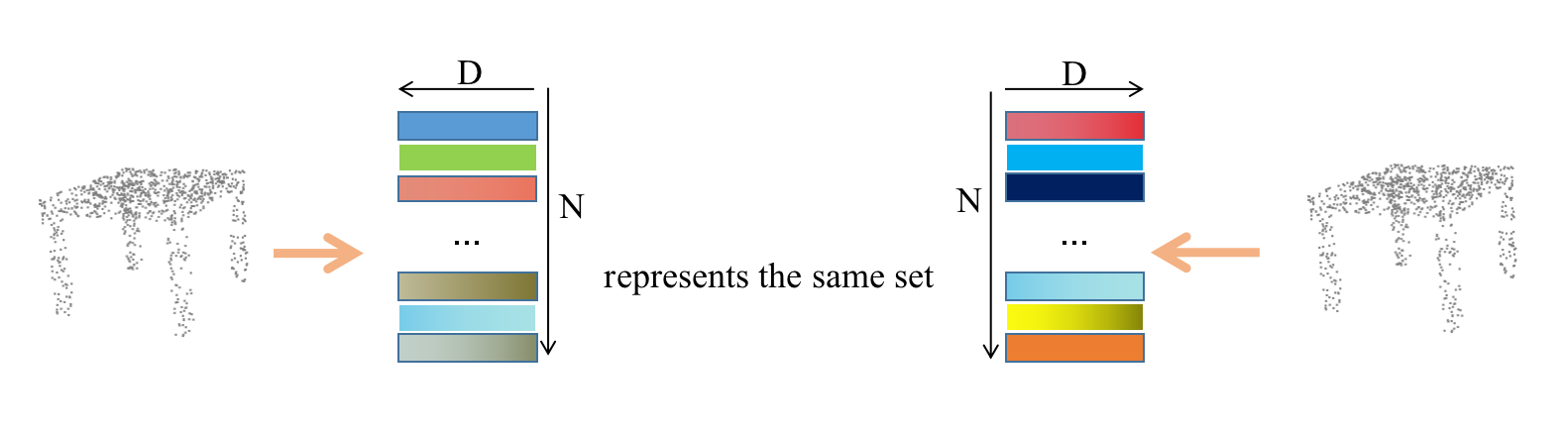}
  \caption{The permutation invariance property of point cloud. Where $N$ denote the numbers of orderless points, $D$ denote the numbers of dimensions of coordinate.}
  \label{fig:pi}
\end{figure}

As point out in \cite{pinot2020randomization}, any deterministic classifier could be outperformed by a randomized one in terms of robustness. We therefore ask, \textbf{can we build a randomized classifier that with consistent clean accuracy?} Motivated by this question, we observed that the 3D point cloud classifier's property can be used to build the randomized classifier. As shown in Figure \ref{fig:pi}, due to the unordered nature of 3D point clouds, most point cloud analysis DNNs are invariant to the order permutation of the input point cloud. In this paper, we utilize this property to transform the input point cloud, and finally build the randomized point cloud classifier. 

The \textbf{main contributions} of this paper are summarized as follows:
\begin{itemize}
    \item To the best of our knowledge, invariant transformations defense (IT-Defense) is the first work that uses the networks' property to break the strongest gradient based attacks. It breaks the attack success rate from 100\% to almost 0\%.
    \item Our IT-Defense have no impact on clean accuracy, which is significantly better than previous defense methods.
    \item IT-Defense is compatible to different DNNs and adversarial defense methods, which can serve as a basic network module for 3D point cloud adversarial defense.
\end{itemize} 

\section{Related Work}
\textbf{Adversarial attacks on images.} The deep neural networks are vulnerable to adversarial examples, it was firstly found in the image domain~\cite{firstadv}. Then a bunch of algorithms to generate adversarial examples or to attack the deep neural networks are proposed~\cite{moosavi2016deepfool,moosavi2017universal,su2019one,physicalaatck}. Which can be summarized as white-box attacks and black-box attacks~\cite{benchmarkingimage}. Most attacks in white-box setting are based on the input gradient. The fast gradient method (FGM)~\cite{fgm} generates adversarial examples by one-step update towards the input gradient. The iterative version of FGM (IFGM)~\cite{ifgm} generate adversarial examples by small steps towards the gradient direction. A momentum term was introduced by ~\cite{mifgm} to stabilize the update direction during the iterations, which was known as MIFGM. The projected gradient descent method (PGD) adopts random starts during the iterations, which was served as a baseline in first-order adversary. The C\&W attack~\cite{cwattack} turn the process of obtaining adversarial examples into an optimization problem, and used the Adam~\cite{adam} for optimization. This algorithm was heavily used in recent point cloud attack research.

\textbf{Adversarial attacks on point clouds.} Due to the safety-critical applications of the point cloud in robotics and safe-driving cars, 3D point clouds has attracted many researchers in the computer vision community. However, its robustness is relatively under-explored compared to the image domain. ~\cite{generatingadpoint} first proposed to generate adversarial point clouds examples via C\&W attack, which introduced two point cloud attacks, point perturbation and adding. The point adding attack can further divide by adding independent points, adding clusters, and adding objects. However, the generate adversarial point clouds examples are very messy, which can be easily perceivable by humans. The knn attack~\cite{knnattack} adopted a kNN distance constraint, a clipping and a projection operation to generate more smooth and imperceptible adversarial point clouds examples. The Geometry-Aware Adversarial Attack ($GeoA^3$) ~\cite{geoa3} further improved the imperceptible to humans. The perturbation based attacks can be removed by statistical outlier removal (SOR) ~\cite{dupnet} method that removes points with a large kNN distance if the perturbation is too large. To solve this problem, ~\cite{attsordefense} developed the JGBA attack, which is an efficient attck to SOR defense. Besides, the point drop attack ~\cite{pointcloudsaliencymaps} was developed by a gradient based saliency map, which iteratively remove the most important points. Moreover, the AdvPC \cite{advpc} improved the transferability of adversarial point cloud examples by utilizing a point cloud auto-encoder, and the LG-GAN \cite{lggan} utilized the powerful GANs \cite{goodfellowgan} to generate adversarial examples guided by the input target labels. However, most of those attacks are integrate the gradient information from the input, which can be a weakness.

\textbf{Adversarial defenses.} To overcome the threat of adversarial examples to DNNs, extensive research has been conducted on defending against adversarial attacks. Existing adversarial defense methods can be roughly divided into two classes: attacking stage defense and testing stage defense. The adversarial training ~\cite{advtfree,youonlypo,fastisfree,cascadetrain} is an effective way to improve the model's robustness, which have two stages defense effect. There are other defense methods that have two stages defense effect. For example, Pang et al.\cite{pang2019rethinkingcrossentropy} proposed a novel loss, the model's adversarial robustness is increased if the model was trained with this loss. Thermometer encoding\cite{thermometerencoding} encode values in a discrete way. EMPIR\cite{sen2020empir} constructed ensemble models with mixed precision of weights and activations. Ensemble diversity\cite{improvingdiversity} improved robustness with a regularization term. For the attacking stage defense, k-Winners Take All\cite{kwiners} developed a novel activation function that masks the backpropagated gradient, input transformations \cite{inputtrans} and input randomization \cite{mitigatingadv} utilized the backpropagated transformed gradient to fool the attacker, stochastic activation\cite{dhillon2018stochastic} replaces the dropout layer with a non-differentiable function. Those defend methods can prevent the attacker from generating adversarial examples effectively.
For the testing stage defense, the are a lot of adversarial examples detection methods\cite{odds,asymetryadversarial,turningweakness,generativeclassifier,lid} are developed, the transformation method of Pixeldefend\cite{song2017pixeldefend}, defense-GAN\cite{defensegan}, and sparse fourier transform \cite{fouriertrans} transforms the adversarial examples to a normal sample, the Mixup Inference\cite{pang2019mixupinference}, ME-Net\cite{me-net}, and Error Correcting Codes\cite{errorcorrecting} mitigate the adversarial perturbations by inference the adversarial examples directly.

We note that Guo et al. \cite{inputtrans} and Xie et al. \cite{mitigatingadv} also utilized input transformation before feeding into the DNN to defend adversarial attacks, which caused the obfuscated gradients effect and can be defeat by the expectation over transformation (EOT) attacks proposed by\cite{obfuscated}. However, our work has several key differences with Guo et al. and Xie et al.. Firstly, we used the invariant transformations of DNN, which do no harm to the DNN, but Guo et al. and Xie et al. have some kind of accuracy drop of the standard model. Secondly, we do not cause information loss between the original sample and the transformed sample. Thirdly, we cannot be defeated by the EOT attack.

\section{Methodology}
\subsection{An overview of point cloud attack}

Let $ {x} \in \mathbf{R}^{N \times 3} $ represents a set of clean 3D points $\{P_i | i = 1, ..., N\}$, and $y$ denote the corresponding true label. For a classifier $F(x) : x \rightarrow y$ that outputs the prediction for an input, the attacker wants to generate an adversarial example $x^{adv}$ which is imperceptible by humans from $x$ but fools the classifier.
We give a brief introduction of some famous attack algorithms on 3D point cloud classifier in this section.

\textbf{FGM} \cite{fgm} generates adversarial example by one-step update.

\begin{equation}
\boldsymbol{x}^{a d v}=\boldsymbol{x}-\epsilon \cdot \operatorname{sign}\left(\nabla_{\boldsymbol{x}} J\left(\boldsymbol{x}, y\right)\right)
\end{equation}
where $\nabla_{\boldsymbol{x}} J$ is the gradient of the loss function with respect to the input $x$. $\operatorname{sign(.)}$ is the sign function that turns the gradient value into the direction value.

\textbf{I-FGM} \cite{ifgm} generate adversarial example by small steps in a iterative fashion.

\begin{equation}
\boldsymbol{x}_{t+1}^{a d v}=\boldsymbol{x}_{t}^{a d v}-\alpha \cdot \operatorname{sign}\left(\nabla_{\boldsymbol{x}} J\left(\boldsymbol{x}_{t}^{a d v}, y\right)\right)
\end{equation}
where $\alpha = \epsilon / T$ with T steps iteration, and ${x}_{0}^{adv}=x$. 

\textbf{MIFGM} \cite{mifgm} introduced a momentum term to stabilize the update direction during the iteration.

\begin{equation}
\boldsymbol{g}_{t+1}=\mu \cdot \boldsymbol{g}_{t}+\frac{\nabla_{\boldsymbol{x}} J\left(\boldsymbol{x}_{t}^{a d v}, y\right)}{\left\|\nabla_{\boldsymbol{x}} J\left(\boldsymbol{x}_{t}^{a d v}, y\right)\right\|_{1}}
\end{equation}

\begin{equation}
\boldsymbol{x}_{t+1}^{a d v}=\boldsymbol{x}_{t}^{a d v}-\alpha \cdot \operatorname{sign}\left(\boldsymbol{g}_{t+1}\right)
\end{equation}
where $g_{t}$ gathers the gradient information up to the $t$-th iteration with a decay factor $\mu$. We mainly compare FGM, I-FGM and MIFGM with GvF-P \cite{fgmpoint}, the first self-robust point cloud defense method.

\textbf{C\&W} \cite{cwattack} turn the process of obtaining adversarial examples into an optimization problem.

\begin{equation}
\underset{\boldsymbol{x}^{a d v}}{\arg \min }\left\|\boldsymbol{x}^{a d v}-\boldsymbol{x}\right\|_{p}+c \cdot J\left(\boldsymbol{x}^{a d v}, y\right)
\end{equation}
where the loss $J$ can be different from the cross-entropy loss and many variants are proposed for point cloud attack\cite{knnattack,generatingadpoint,geoa3}.

\subsection{Invariant transformation defense (IT-Defense)}

\begin{figure*}
  \includegraphics[width=\linewidth]{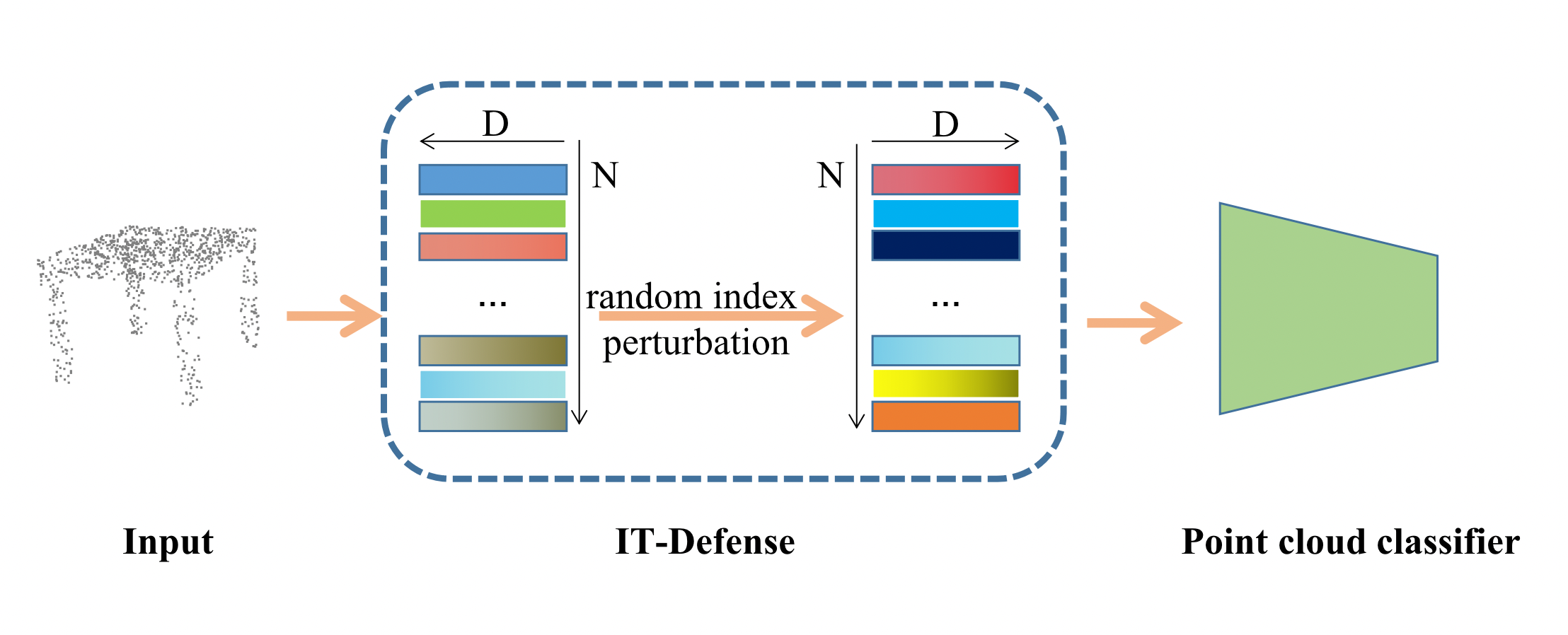}
  \caption{The pipeline of our proposed IT-Defense.}
  \label{fig:pipeline}
\end{figure*}

In this paper, we make full use of the nature of point cloud classifiers, i.e., the point cloud classifiers are permutation invariant to the input point cloud's index, and propose the invariant transformation defense (IT-Defense).
The pipeline of IT-Defense is shown in Figure \ref{fig:pipeline}.
The IT-Defense can be described as following: 
\begin{equation}
g(x)={\nabla_{t(\boldsymbol{x})}} J\left(t(\boldsymbol{x}), y\right)
\end{equation}
where $\nabla_{t(\boldsymbol{x})} J$ is the gradient of the loss function with respect to the random input transformation $t()$ of $x$. Note that unlike previous random input transformation defense \cite{inputtrans,mitigatingadv,raff2019barrage}, our transformation is invariant to the classier $F()$.
The intuition is that by randomly perturb the index of the input point cloud, the transformed point cloud will not affect the performance of point cloud classifier due to the orderless nature of point cloud, but the real gradient have perturbed significantly, thus fool the attacker.

Following, we perform some simple analysis to investigate why IT-Defense works. 

\subsection{Theoretical analysis of IT-Defense against  gradient-based  attacker}\label{unified}
In the following section, we use the theory of \cite{ren2021towardsunified} to analyze IT-Defense against various attacks. Under the system of game theory, the relationship between the adversarial robustness of neural networks and the complexity of game interaction is modeled uniformly. They proved that the adversarial perturbations mainly affect high-order interactions. They define the game-theoretic interactions as follows:

\begin{equation}
I(i, j)=\tilde{\phi}(i \mid N)_{j \text { always present }}-\tilde{\phi}(i \mid N)_{j \text { always absent }}
\end{equation}
where $\tilde{\phi}(i \mid N)_{j \text { always present }}$ denote the importance of input variable $i$ when $j$ is always present and $\tilde{\phi}(i \mid N)_{j \text { always absent }}$ denote the importance of input variable $i$ when $j$ is always absent. Then, the interactions can be decomposed into multiple orders \cite{zhang2020interpretingrobustness} as following:
\begin{equation}
I(i, j)=\frac{1}{n-1} \sum_{m=0}^{n-2} I_{i j}^{(m)}, \quad I_{i j}^{(m)}=\mathbb{E}_{S \subseteq N \backslash\{i, j\}}[\Delta v(i, j, S)]
\end{equation}
where $I_{i j}^{(m)}$ represents the $m$th-order interaction. $m$ represents the number of units in the background $S$ other than input units $i$ and $j$, reflecting the complexity expressed by the interaction. When the background contains more input units, $m$ is larger, and the game-theoretic interaction between $i$ and $j$ can be regarded as a high-order interaction. When the background contains a small number of input units, $m$ is small, the game-theoretic interaction between $i$ and $j$ is a low-order interaction.

\begin{figure}[ht]
  \includegraphics[width=\linewidth]{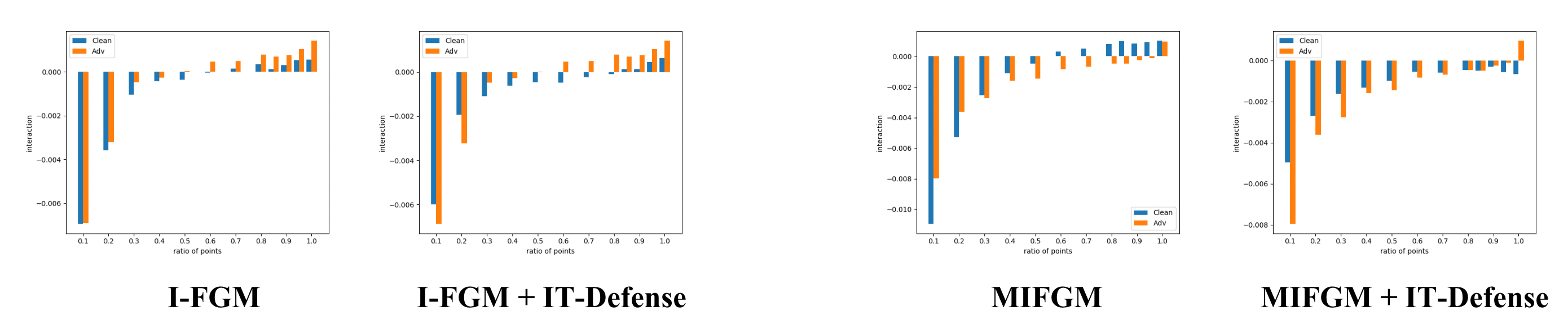}
  \caption{The multi-order interaction $I_{clean}^{(m)}$ and $I_{adv}^{(m)}$ of gradient-based attacks. Our IT-Defense mainly affect the low-order interactions, thus mitigate adversarial effects.}
  \label{gradient}
\end{figure}

As shown in Figure \ref{gradient}, we perform I-FGM and MIFGM and IT-Defense against their in Pointnet \cite{pointnet}, where the lower ratio of points represent the the low-order interactions, and higher ratio of points represent the the high-order interactions, the results show that the adversarial example without defense mainly affected high-order interactions, the IT-Defense mainly change the low-order interactions, thus mitigate adversarial effects. 

\subsection{Theoretical analysis of IT-Defense against optimization-based attacker}
\begin{figure}[ht]
  \includegraphics[width=\linewidth]{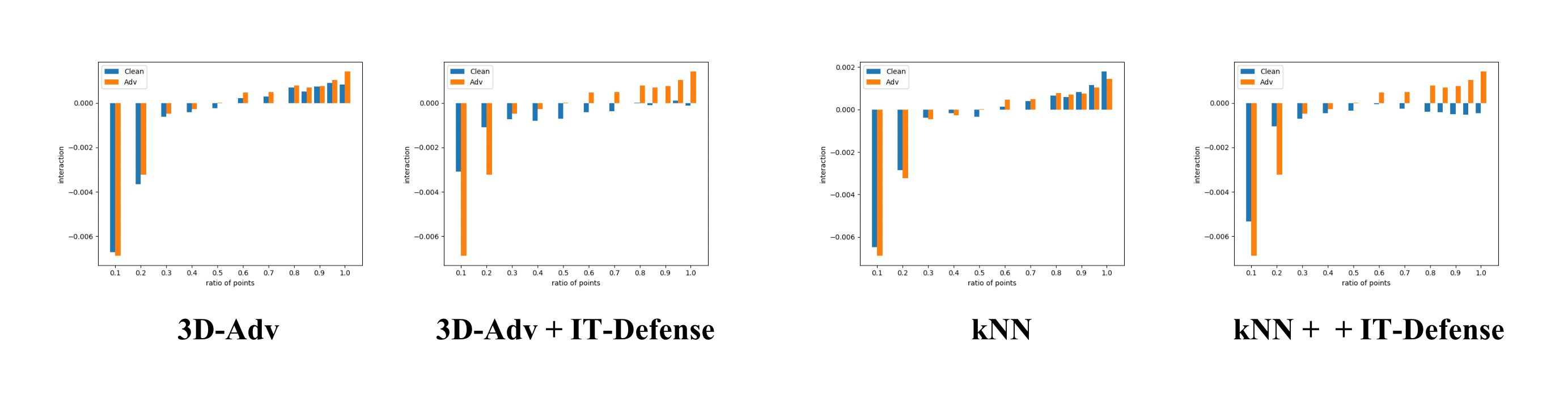}
  \caption{The multi-order interaction $I_{clean}^{(m)}$ and $I_{adv}^{(m)}$ of optimization-based attacks. Our IT-Defense mainly affect the low-order interactions, thus mitigate adversarial effects.}
  \label{optimize}
\end{figure}

Following Sec. \ref{unified}, two optimization-based attacker are selected, the 3D-Adv \cite{generatingadpoint} and kNN attack \cite{knnattack}. From Figure \ref{optimize}, we can observe similar results as Sec. \ref{unified}, which means the IT-Defense mainly change the low-order interactions to overcome adversarial effects.

\subsection{Theoretical analysis of IT-Defense against adaptive attacks}

Recent works on robust defense \cite{tramer2020adaptive, obfuscated, gradmasking} suggest that the proposed new defense algorithms should take a further evaluation against the corresponding adaptive attack.
Since our method transform the index the points, the expectation over transformation (EOT) attacks proposed by\cite{obfuscated} is expect as the adaptive attacker, due to its excellent performance in adaptive attack to the input transform based methods \cite{raff2019barrage}. 
Following ~\cite{raff2019barrage}, we use the EOT to build a strong attacker. The EOT \cite{eot} is defined as follows:
\begin{figure}[h]
  \includegraphics[width=\linewidth]{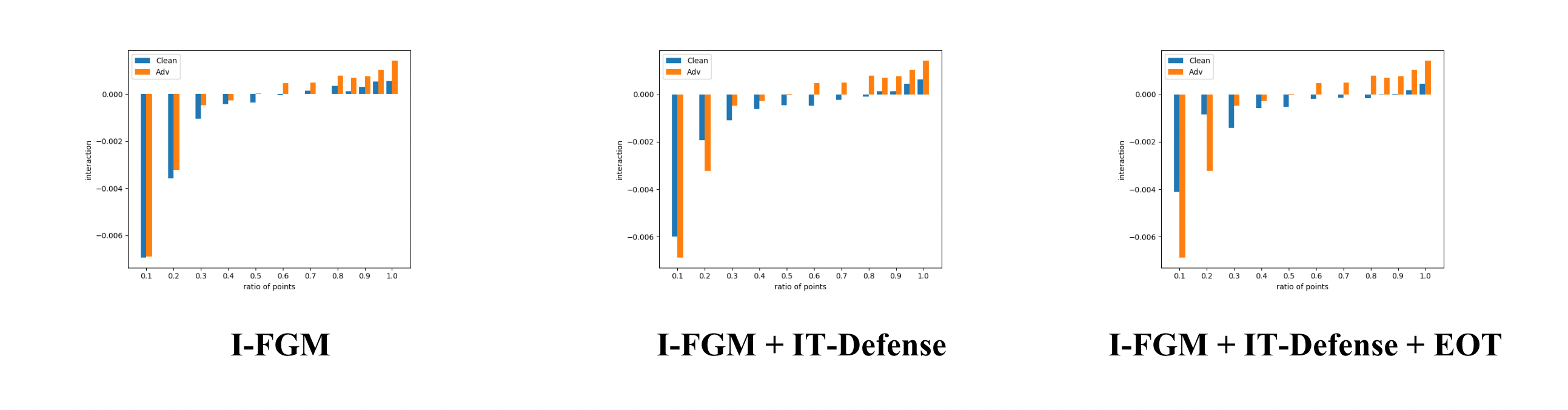}
  \caption{The multi-order interaction $I_{clean}^{(m)}$ and $I_{adv}^{(m)}$ of Pointnet and Pointnet with our proposed IT-Defense, and IT-Defense with EOT attack. The EOT attack further increased the gap in low-order interactions, thus not effective to attack the IT-Defense.}
  \label{eot_theory}
\end{figure}
\begin{equation}
\nabla_{x} \mathbb{E}_{r}\left[f_{r}(x)\right]=\mathbb{E}_{r}\left[\nabla_{x} f_{r}(x)\right] \approx \frac{1}{n} \sum_{i=1}^{n} \nabla_{x} f_{r_{i}}(x)
\end{equation}
where $f_{r}(x)$ is the randomized classifier, and the $r_{i}$ are independent draws of the randomness transformation. 
But our transformation is big enough (up to $1024!$ or $2048!$ depends on $n$), given one transformation $t(.)$ of our defender, the probability of correctly sample this $t(.)$ by an attacker is $1/{n!}$, which is sufficently small, simply repeat the transformation multiple times (usually 10 to 30 times in most EOT attacks literatures~\cite{tramer2020adaptive}) cannot recover the true gradient.
As shown in Figure \ref{eot_theory}, the EOT attack further increased the gap in low-order
interactions compare to without EOT attack, thus it is not effective to attack the IT-Defense.
We further verified this by experiment.
\section{Experiments}
In this section, we present the experimental results to demonstrate the effectiveness of the proposed method. We first specify the experimental settings. Then we conduct extensive experiments to study the defense effects of IT-Defense. And perform comparative experiments to analyze IT-Defense in detail.

\subsection{Experimental Settings}
We implement all experiments on a Linux server with 8 Nvidia RTX 3090 GPUs. For point cloud attacks, we use the ModelNet40~\cite{modelnet40} test set kindly provided by\cite{ifdefense}, which also contained the targeted label for targeted attacks. We also select three commonly used networks in 3D computer vision area~\cite{pointcutmix,isometry3d} for evaluation, \ie, PointNet~\cite{pointnet}, PointNet++~\cite{pointnet++}, and DGCNN~\cite{dgcnn}, the SOTA networks such as PointConv \cite{pointconv}, PAConv \cite{xu2021paconv}, Point Cloud Transformer (PCT) \cite{guo2021pct} and CurveNet \cite{xiang2021walk} are also selected for KNN attack evaluation. The FGM\cite{fgm}, I-FGM\cite{ifgm}, MIFGM\cite{mifgm} and PGD\cite{pgd} attacks were selected as the gray-box attacker. We follow the same settings as \cite{fgmpoint} for point cloud attack. Moreover, the untargeted point dropping attack~\cite{pointcloudsaliencymaps}, and the C\&W attack's\cite{cwattack} variant kNN attack~\cite{knnattack} and 3D-Adv attack~\cite{generatingadpoint} were further verified our effective defense ability. The JGBA attack and $GeoA^3$ attack were implemented by their open-sourced code and dataset. 
Experiments are repeated 5 times. The estimated gradient of EOT is averaged by 10 random transformations.

\subsection{Results}
\subsubsection{IT-Defense does not reduce clean accuracy.}
IT-Defense only change the order of points, it does not reduce clean accuracy. In this section, we validate this via experiments. As shown in Table \ref{cleanacc}, unlike previous SOTA defenses that reduce the clean accuracy, such as the Simple Random Sampling~\cite{yang2019adversarial} (SRS), Statistical Outlier Removal~\cite{yu2018pu} (SOR), DUP-Net~\cite{dupnet} and the IF-Defense~\cite{ifdefense} are reduced the clean accuracy up to 4\%, the IT-Defense that build upon the property of point cloud recognition model does not reduce clean accuracy. This is an important property, which means that IT-Defense can be embedded in any point cloud recognition model.


\begin{table*}[t]
\centering
\setlength{\tabcolsep}{3pt}
\renewcommand\arraystretch{1.15}
\caption{Classification accuracy of various defense methods on clean ModelNet40 by Pointnet. The best results for each row are emphasized as bold.}
\label{cleanacc}
\vspace{2mm}
\begin{tabular}{ c c c c c c c}
\toprule
Model & Clean & SRS~\cite{yang2019adversarial} & SOR~\cite{yu2018pu} & DUP-Net~\cite{dupnet} & IF-Defense~\cite{ifdefense} & Ours \\
\midrule

PointNet & 88.49 & 87.24 & 87.80 & 86.83 & 84.20 & \bf88.49

\\
\bottomrule
\end{tabular}
\end{table*}

\subsubsection{IT-Defense against various attackers.}
In this section, we show the experimental results of the proposed invariant transformation defense (IT-Defense) method with different attackers. We firstly perform the classical adversarial attacks to our IT-Defense, the results are shown in Table \ref{whiteaatck}. We report the success rates of FGM, I-FGM, I-FGM+EOT, PGD, and MIFGM attacks against our defense method and in no defense setting. The I-FGM+EOT attack cannot break our defense and getting worse results than I-FGM attack.
The results shown that our defense improve the robustness of the model against various attacks significantly, our method break the success rate from 98.87 to 0.45 in some case.

\begin{table*}[h]
\begin{center}
\centering
\caption{The success rates (\%) of targeted attacks.$*$ denotes that results are reported in GvG-P~\cite{fgmpoint}.}
\label{whiteaatck}
\footnotesize
\begin{tabular}{c|c|c|c|c|c|c}
\toprule
Model& Attack & FGM & I-FGM & I-FGM+EOT & PGD & MIFGM \\
\hline
\multirow{2}{*}{Pointnet} & No Defense& 3.69 & 98.87 & 98.87 & 98.78 & 85.29  \\
                        & IT-Defense & \bf0.64$\pm 0.05$ & \bf0.49$\pm 0.06$ & \bf0.20$\pm 0.04$ & \bf0.93$\pm 0.05$ & \bf0.47$\pm 0.04$ \\
\hline
\multirow{3}{*}{Pointnet++} & No Defense & 2.96 & 92.63 & 92.63 & 93.40 & 12.48 \\
                        & GvG-P* & 3.20 & 69.00 & - & 69.41 & 37.88 \\
                        & IT-Defense & 3.03$\pm 0.07$ & \bf1.27$\pm 0.16$ & \bf0.23$\pm 0.10$ & \bf1.99$\pm 0.21$ & \bf1.21$\pm 0.11$\\
\hline
\multirow{2}{*}{DGCNN} & No Defense & 3.36 & 78.65 & 78.65 & 78.00 & 23.34\\
                        & IT-Defense & \bf3.26$\pm 0.05$ & \bf1.03$\pm 0.04$ & \bf0.32$\pm 0.04$ & \bf1.80$\pm 0.12$ & \bf1.07$\pm 0.09$\\
\bottomrule
\end{tabular}
\end{center}
\vspace{-1ex}
\end{table*}



The point dropping attack~\cite{pointcloudsaliencymaps} is based on the saliency maps of the input point cloud, where also used the gradient in some degree, we therefore perform this attack to our defense. For fairness, we used the same settings with \cite{pointcutmix,ifdefense} and used classification accuracy to compare with state of the art defense algorithms, the results are shown in Table \ref{dropattack}. We can see that for Drop 200 and Drop 100 attacks, the IT-defense leads to better results than the IF-Defense. The results verified that our defense method can be applied to any kind of attack based on gradient information.

\begin{table*}[h]
\centering
\setlength{\tabcolsep}{3pt}
\renewcommand\arraystretch{1.15}
\caption{Classification accuracy of various defense methods on ModelNet40 under point dropping attack~\cite{pointcloudsaliencymaps}. Drop 200 and Drop 100 denote the dropping points is 200 and 100 respectively. $*$ denotes that results are reported in IF-Defense~\cite{ifdefense}. We report the best result of three IF-Defense. The best results for each row are emphasized as bold.}
\label{dropattack}
\vspace{2mm}
\begin{tabular}{c c c c c c c c}
\toprule
Attack & Model & No Defense$*$ & SRS$*$ & SOR$*$ & DUP-Net$*$ & IF-Defense$*$ & Ours \\
\midrule
\multirow{2}{*}{Drop 200} & 
PointNet & 40.24 & 39.51 & 42.59 & 46.92 & 66.94 & \bf88.02$\pm 0.19$  
\\
\cline{2-8}~
&PointNet++ & 68.96 & 39.63& 69.17 & 72.00 & 79.09 & \textbf{86.83}$\pm 0.49$
\\
\cline{2-8}~
& DGCNN & 55.06 & 63.57& 59.36 & 36.02 & 73.30 & \textbf{83.01}$\pm 0.15$
\\
\midrule
\multirow{3}*{Drop 100} & 
PointNet & 64.67 & 63.57 & 64.75 & 67.30 & 77.76 & \bf88.33$\pm 0.19$  
\\
\cline{2-8}~
&PointNet++ & 80.19 & 64.51& 74.16 & 76.38 & 84.56 & \textbf{88.31}$\pm 0.16$ 
\\
\cline{2-8}~ &DGCNN & 75.16 & 49.23& 64.68 & 44.45 & 83.43 & \textbf{87.86}$\pm 0.10$
\\
\bottomrule
\end{tabular}
\end{table*}

We further verified our defense on kNN attack~\cite{knnattack}, 3D-Adv attack~\cite{generatingadpoint}, JGBA attack\cite{attsordefense} and $GeoA^3$ attack\cite{geoa3}, three of them are the C\&W attack's\cite{cwattack} variant in point cloud. The C\&W attack turns the process of obtaining adversarial examples into an optimization problem, but the gradient information is needed in every iteration step within the optimization process. We report the success rates of the kNN attack~\cite{knnattack}, the 3D-Adv attack~\cite{generatingadpoint}, the JGBA attack\cite{attsordefense} and the $GeoA^3$ attack\cite{geoa3} against IT-Defense in Table \ref{knn_gen}. In general, the IT-Defense consistently breaks the success rate from a high level (near 100\%) to near 0\%, which means that we can totally break the attack effect caused by the attacker.


\begin{table*}[ht!]
\centering
\setlength{\tabcolsep}{3pt}
\renewcommand\arraystretch{1.15}
\caption{The success rates (\%) of targeted attacks of kNN attack~\cite{knnattack}, 3D-Adv attack~\cite{generatingadpoint}, JGBA attack\cite{attsordefense} and the $GeoA^3$ attack\cite{geoa3}. The best results for each row is emphasized as bold.}
\label{knn_gen}
\vspace{2mm}
\begin{tabular}{c c c c}
\toprule
Attack & Model & No Defense & IT-Defense\\
\midrule
\multirow{7}{*}{kNN} &
PointNet & 85.45& \textbf{0.41}
\\
& PointNet++ & 99.96&\textbf{0.51}
\\
&DGCNN &60.53 &  \textbf{0.69}
\\
&PointConv \cite{pointconv}&89.75 &  \textbf{0.61}
\\
&PAConv \cite{xu2021paconv}&99.96 &  \textbf{2.76}
\\
&PCT \cite{guo2021pct}&98.78 &  \textbf{0.59}
\\
&CurveNet \cite{xiang2021walk} &85.53 &  \textbf{0.36}
\\
\midrule
\multirow{3}*{3D-Adv} &
PointNet &100.00 & \textbf{0.20}
\\
&PointNet++ &100.00 &\textbf{0.61}
\\
&DGCNN &100.00 & \textbf{0.36}
\\
\midrule
JGBA &
PointNet &100.00 & \textbf{0.19}
\\
\midrule
$GeoA^3$ &
PointNet &100.00 & \textbf{32.00}
\\

\bottomrule
\end{tabular}
\end{table*}




\subsection{Comparative experiments}
In this section, we explore different attack settings on IT-Defense.
\subsubsection{Influence of perturbation budget}
As suggest in \cite{dong2020benchmarking}, the perturbation budget have significant impact on the attack performance. We therefore perform experiment with different perturbation budget in the range of $[0.05,0.40]$, with point cloud in $[0,1]$. As shown in Figure \ref{comparative}, our IT-Defense is robust within the $0.25$ perturbation budget, it's big enough to be easily detected by humans.
\subsubsection{Influence of attack steps}
\begin{figure}[ht]
  \includegraphics[width=\linewidth]{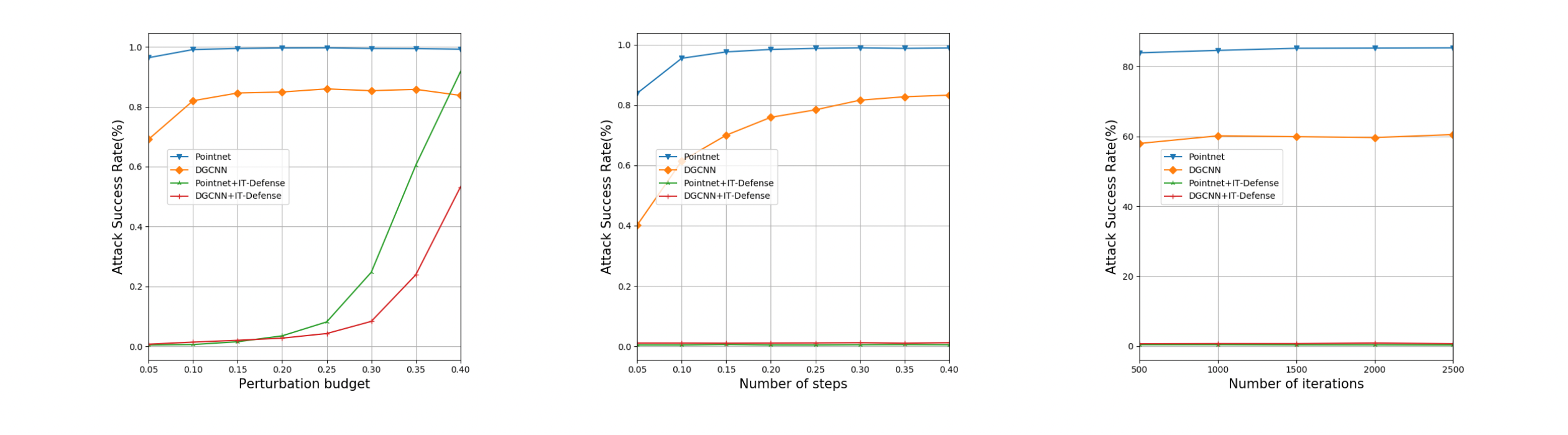}
  \caption{The Attack Success Rate(\%) vs perturbation budget and number of steps for IFGM attack, the Attack Success Rate(\%) vs number of iterations for kNN attack.}
  \label{comparative}
\end{figure}

The number of attack steps of gradient-based attack is another important factor that affect the attack performance. The results are shown in Figure \ref{comparative}, for Pointnet and DGCNN, the Attack Success Rate(\%) increased with the number of attack steps for IFGM without our IT-Defense, but maintain almost constant zero Attack Success Rate(\%) with IT-Defense.

\subsubsection{Influence of attack iterations}
The number of iterations of optimization-based attack is a vital variable during attack. We perform KNN attack \cite{knnattack} on Pointnet and DGCNN with the iterations within $[500, 2500]$. As shown in Figure \ref{comparative}, for Pointnet and DGCNN, similar to the number of steps for IFGM attack, the Attack Success Rate(\%) increased slightly with the number of attack iterations for KNN attack without our IT-Defense, but maintain almost constant zero Attack Success Rate(\%) with IT-Defense.

\subsubsection{Influence of EOTs}
In order to validate our IT-Defense resistent to EOTs, we scale the number of EOTs up to 100. To the best our knowledge, this is the largest EOT have been performed. The results are shown in Table \ref{eots}, the Attack Success Rate(\%) increased slightly with the number of EOTs initially, but decreased as the EOTs bigger than 50, indicating that EOT attack is not effective to attack our IT-Defense.

\begin{figure*}[htp!]
  \includegraphics[width=\linewidth]{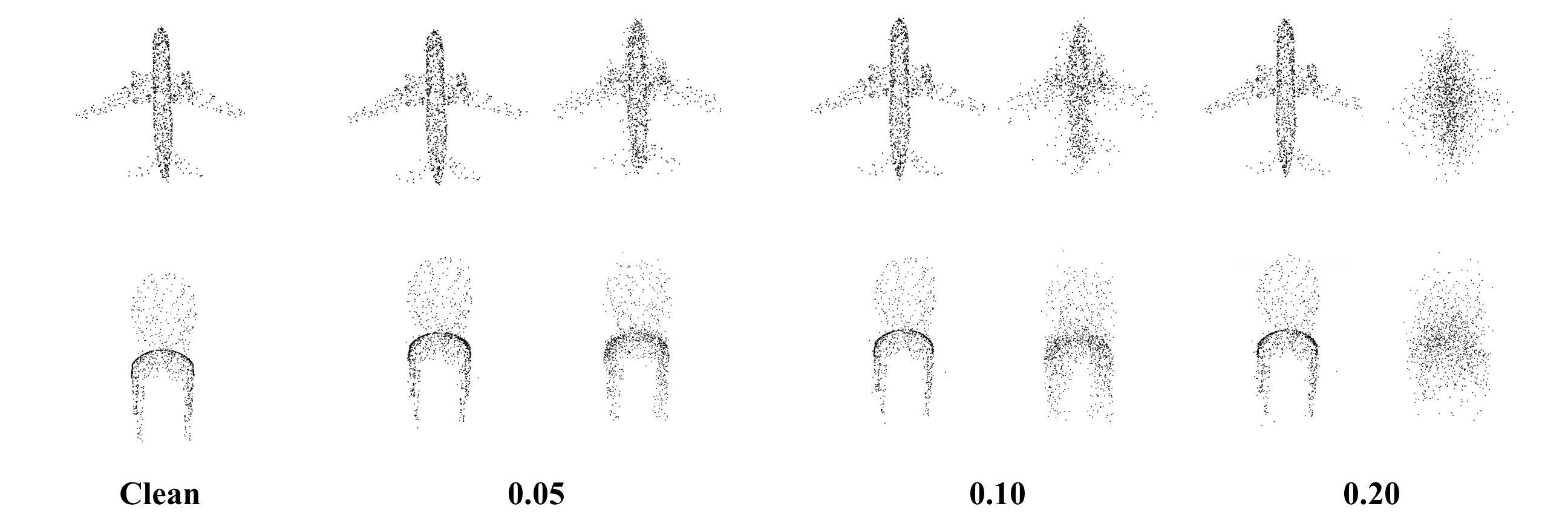}
  \caption{The visualization of adversarial point cloud samples with different perturbation budget. For each group within the same budget, the left and right are adversarial point cloud samples without and with IT-Defense, respectively.}
  \label{pl_fig}
\end{figure*}

\begin{table*}[hpt!]
\centering
\setlength{\tabcolsep}{3pt}
\renewcommand\arraystretch{1.15}
\caption{The success rates (\%) of IT-Defense against EOT attack.}
\label{eots}
\vspace{2mm}
\begin{tabular}{ c c c c c c c c c c c}
\toprule
EOTs & 10 & 20 & 30 & 40 & 50 & 60 & 70 & 80 & 90 & 100\\
\midrule

Success rates & 0.20 & 0.41 & 0.41 & 0.41 & 0.32 & 0.32 & 0.28 & 0.28 & 0.24 & 0.24

\\
\bottomrule
\end{tabular}
\end{table*}

\subsubsection{Understanding the IT-Defense}
To figure out what IT-Defense brings to the adversarial point cloud examples, we visualised the airplane and chair of clean sample and adversarial samples without and with IT-Defense under different perturbation budget. From Figure \ref{pl_fig}, we can conclude that IT-Defense can help models escape the 'adversarial region' when the perturbation budget is small, and making adversarial point cloud examples human recognizable when the perturbation budget is large.

\section{Conclusion}\label{sec:Conclusion}

In this paper, we propose a defense strategy that uses the networks' property to break the adversarial attacks.
Our findings are insightful, the network's property is utilized to defend against attacks, and the results show that our defense can break most of the existing point cloud attacks. 
It is worth mentioning that, although IT-Defense was shown to be a powerful defense mechanism
against adversarial attacks, one limitation might come across, IT-Defense require the deep neural networks has some invariant transformations of input.
Note that our method can only resist attacks for finding adversarial samples, but the adversarial samples generated from the original model can transferred well to IT-Defense. However, our method can easily incorporate with the model with higher robustness, such as models trained with ensemble adversarial training\cite{ensembleadvt} and PointCutMix\cite{pointcutmix}, and combined with other defense methods, thus not only making the attacker hard to generate effective adversarial examples, but also letting our AI system more robust resist to adversarial samples generated by other models.

\section{Acknowledgments}
The project was supported by Innovation Project of Guangxi Graduate Education (YCBZ2021019).

\bibliography{ref/ref}

\vspace{-30pt}
\begin{IEEEbiography}[{\includegraphics[width=1in,height=1.25in,clip,keepaspectratio]{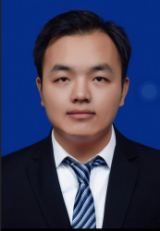}}]{Jinlai Zhang} received the B.S. degree from Changsha University of Science and Technology, Changsha, China, in 2017. He is currently pursuing the Ph.D. with the College of Mechanical Engineering, Guangxi University. His research interests include 3D deep learning and time series forecasting.
\end{IEEEbiography}
\vspace{-30pt}

\begin{IEEEbiography}[{\includegraphics[width=1in,height=1.25in,clip,keepaspectratio]{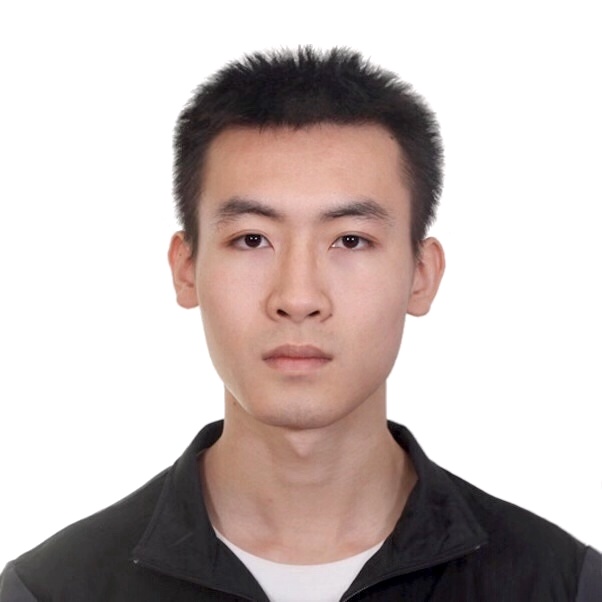}}]{Yinpeng Dong} received his BS and PhD de- grees from the Department of Computer Science and Technology in Tsinghua University, where he is currently a postdoctoral researcher. His research interests are primarily on the adver- sarial robustness of machine learning and deep learning. He received Microsoft Research Asia Fellowship and Baidu Fellowship.
\end{IEEEbiography}
\vspace{-30pt}

\begin{IEEEbiography}[{\includegraphics[width=1in,height=1.25in,clip,keepaspectratio]{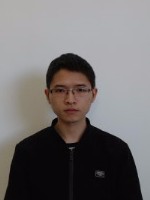}}]{Binbin Liu} received the B.S. degree from University of Science and Technology of China, in 2019. He is currently pursuing the Master with the Department of Computer Science and Technology, Tsinghua University. His research interests include 3D deep learning and robustness of object detection.
\end{IEEEbiography}
\vspace{-30pt}

\begin{IEEEbiography}[{\includegraphics[width=1in,height=1.25in,clip,keepaspectratio]{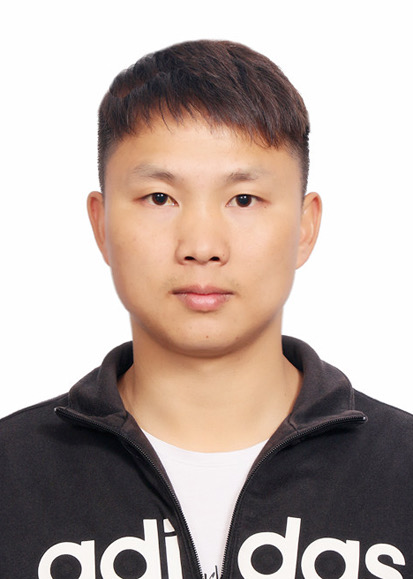}}] {Bo Ouyang} received B.E. degree from National University of Defense Technology in 2011. He is currently working towards the MA. degree in the Department of Computer Science and Technology from Tsinghua University. His research interests include Computer Vision, Deep Learning, and Natural Language Processing.
\end{IEEEbiography}
\vspace{-30pt}


\begin{IEEEbiography}[{\includegraphics[width=1in,height=1.25in,clip,keepaspectratio]{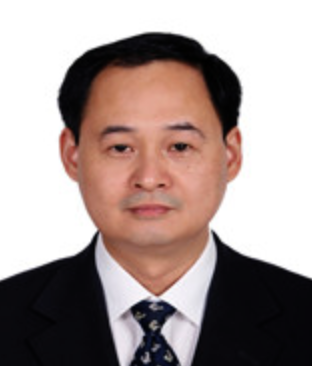}}]{Jihong Zhu}  received his Ph.D. degree from Nanjing University of Science and Technology, Nanjing, in 1995. Currently, he is a professor with the Department of Precision Instrument,Tsinghua University, Beijing, China. His research interests include flight control and robotics.
\end{IEEEbiography}
\vspace{-30pt}

\begin{IEEEbiography}[{\includegraphics[width=1in,height=1.25in,clip,keepaspectratio]{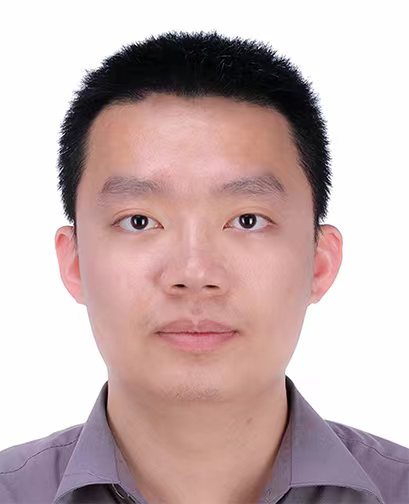}}]{Minchi Kuang}  received his doctorate from Tsinghua University in 2017. He is now working as an assistant researcher in the Department of precision instruments of Tsinghua University. His main research interests are artificial intelligence and intelligent control.
\end{IEEEbiography}
\vspace{-30pt}

\begin{IEEEbiography}[{\includegraphics[width=1in,height=1.25in,clip,keepaspectratio]{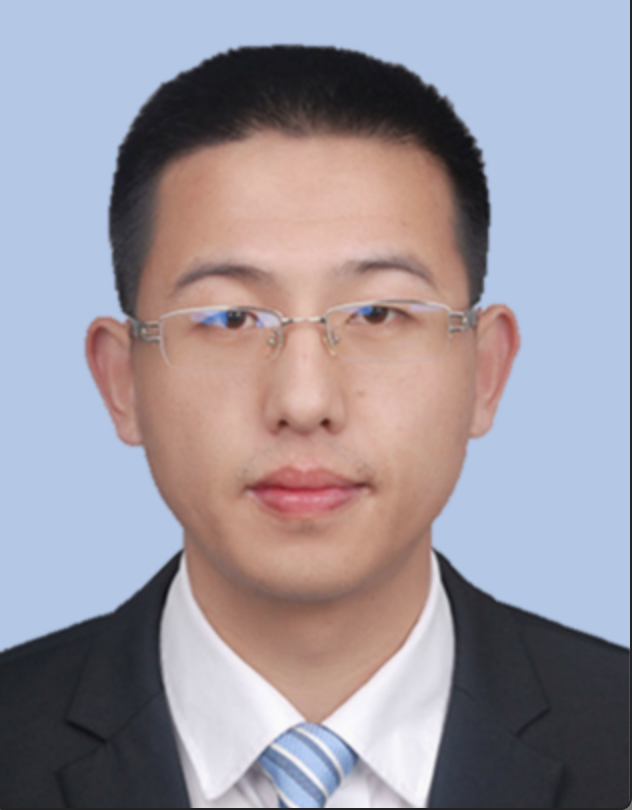}}]{Houqing Wang}  received the B.S. degree in ship electrical and electronic engineering from the Shanghai Maritime University in 2015, and M.S. and Ph.D. degrees both in power electronics and power drives from the Shanghai Maritime University, Shanghai, China, in 2017 and 2020, respectively.
From April to September 2019, he was a Research Assistant with the Centre for Smart Energy Conversion and Utilization Research in the City University of Hong Kong. Since January 2021, he has been with the Tsinghua University, Beijing, China, where he is currently a Postdoctor in the Department of Precision Instrument. His research interests include digital control techniques of power converters, renewable energy generation systems, and battery energy storage systems.
\end{IEEEbiography}
\vspace{-30pt}


\begin{IEEEbiography}[{\includegraphics[width=1in,height=1.25in,clip,keepaspectratio]{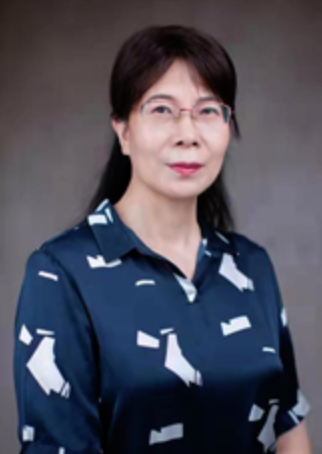}}]{Yanmei Meng} received the Ph.D. degree from Hefei University of Technology, Hefei, China, in 2003, respectively. She is currently a Professor with the College of Mechanical Engineering, Guangxi University. Her research interest includes intelligent detection and control technology, computer vision.
\end{IEEEbiography}
\vspace{-30pt}

\end{document}